\def\year{2022}\relax
\documentclass[letterpaper]{article} 
\usepackage{aaai22}  
\usepackage{times}  
\usepackage{helvet}  
\usepackage{courier}  
\usepackage[hyphens]{url}  
\usepackage{graphicx} 
\usepackage{natbib}  
\usepackage{caption} 
\DeclareCaptionStyle{ruled}{labelfont=normalfont,labelsep=colon,strut=off} 
\usepackage{algorithm}
\usepackage{algorithmic}
\usepackage{subcaption}

\usepackage{xcolor}

\usepackage{newfloat}
\usepackage{listings}
\floatstyle{ruled}
\newfloat{listing}{tb}{lst}{}
\floatname{listing}{Listing}
\title{Reinforcement Learning for Block Decomposition of CAD Models}
\author{
    Benjamin C. DiPrete\textsuperscript{\rm 1,2},
    Rao V. Garimella\textsuperscript{{\rm 2}}\thanks{Corresponding author},
    Cristina Garcia Cardona\textsuperscript{{\rm 2}}, Navamita Ray\textsuperscript{{\rm 2}} 
}
\affiliations{
    \textsuperscript{\rm 1}Georgia Institute of Technology, Atlanta, Georgia USA (bendiprete@gmail.com)    

   \textsuperscript{\rm 2}Los Alamos National Laboratory, PO Box 1663, Los Alamos, NM 87545,
    (rao@lanl.gov, cgarciac@lanl.gov, nray@lanl.gov)
}

\usepackage{bibentry}

\begin{document}

\maketitle

\begin{abstract}
We present a novel AI-assisted method for decomposing (segmenting) planar CAD (computer-aided design) models into well shaped rectangular blocks as a proof-of-principle of a general decomposition method applicable to complex 2D and 3D CAD models. The decomposed blocks are required for generating good quality meshes (tilings of quadrilaterals or hexahedra) suitable for numerical simulations of physical systems governed by conservation laws. The problem of hexahedral mesh generation of general CAD models has vexed researchers for over 3 decades and analysts often spend more than 50\% of the design-analysis cycle time decomposing complex models into simpler parts meshable by existing techniques. Our method uses reinforcement learning to train an agent to perform a series of optimal cuts on the CAD model that result in a good quality block decomposition. We show that the agent quickly learns an effective strategy for picking the location and direction of the cuts and maximizing its rewards as opposed to making random cuts.  This paper is the first successful demonstration of an agent autonomously learning how to perform this block decomposition task effectively thereby holding the promise of a viable method to automate this challenging process.
\end{abstract}

\section{Introduction}
Many numerical methods compute approximate solutions over a mesh of topologically simpler elements (tetrahedra, hexahedra) representing the computational domain. In highly non-linear problems (e.g. fluid dynamics with shocks), hexahedra are preferred, or even required, over tetrahedra because of their superior accuracy and directional control of the solution \cite{Wang04}. In spite of 30+ years of research, however, there are no reliable algorithms that can automatically generate hexahedral meshes for general CAD models. Contrast this with tetrahedral meshing which has long been automatic at scale for realistic industrial problems \cite{Shephard91, Owen16}.

\begin{figure}[!h]
  \begin{center}
    \includegraphics[width=\columnwidth]{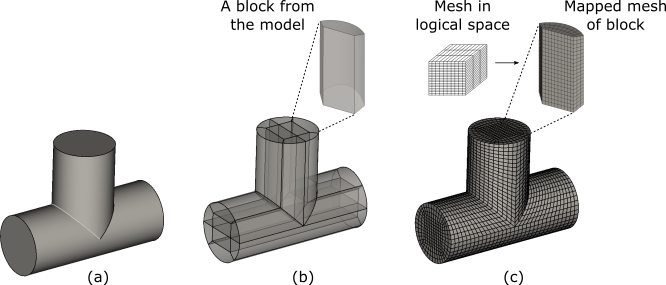}
    \caption{(a) CAD model that cannot be automatically meshed (b)
      model is subdivided into 6-sided blocks using cuts (c)
      each block is meshed by mapping a regular hexahedral
      mesh of a unit cube onto the block.}
    \label{fig:two-pipe-example}
  \end{center}
\end{figure}

To mesh a realistic 3D CAD model (also called a \textit{geometric model}\footnotemark) with hexahedra, analysts use geometric modeling operations (e.g. cutting the domain with a surface) to manually segment the model into a set of simpler shapes or ``blocks'' that can be mapped to a trivially meshable canonical cube \cite{Sarrate14} (Figure~\ref{fig:two-pipe-example}). This process, called \textit{Block Decomposition}, is guided by human intuition and acquired domain expertise that readily ''sees'' how to subdivide a model for a particular application. Attempts to automate this process have not proven generalizable to arbitrary shapes \cite{White04, wang2017sheet}.\footnotetext{In general we will use the word model or shape to implictly mean a geometric model and explicitly point out when we use it to mean a machine learned approximation of reality.}

\subsection{Previous Work}

There have been sustained efforts \cite{Owen16} to develop automatic algorithms to generate hexahedral meshes for complex geometric models since 3D Finite Element Methods became popular. An elementary method (ca. 1980) called \textit{mapped meshing} uses transfinite interpolation to map the structured mesh of a canonical cube to topologically equivalent but geometrically different domains \cite{Gordon82}. For roughly tubular geometric models, an algorithm called \textit{multi-sweeping} \cite{Mingwu96, Shepherd00} extrudes a quadrilateral mesh on one set of faces to form stacks of elements that reach an opposite set.

The \textit{Block Decomposition} method targeted here generalizes these techniques by decomposing complex geometric models into parts that are amenable to mapped meshing or multi-sweeping. Block decomposition is favored by seasoned analysts for its superior control of mesh quality and directionality despite the fact that it must be done manually. While there have been significant efforts to devise automatic decomposition algorithms of complex models based on the model characteristics directly \cite{Lu99, White04, White00, Lu17, wang2017sheet} or on alternate model representations like the medial-axis transform \cite{Blum67,Price95}, most methods have remained experimental or work on a limited class of problems.

In recent years, there has been an sharp uptick in research into using artificial intelligence (AI) or machine learning (ML) with deep neural networks (NN) for solving meshing related problems. Most of the work, however, has focused on using AI/ML for generating or tweaking 2D triangular meshes with point densities suited for a particular PDE (partial differential equation) solution bypassing mesh adaptation using \textit{a posteriori} error estimation \cite{Zhang20, Huang21, Dielen21, Pak21, Bohn21, Yang22, Wu22}. Only Pan et.al. \cite{Pan21, Pan22} describe an actual ML-based quadrilateral mesh generation method but they state that their method only handles simply-connected 2D domains and is very slow, making it unlikely to scale to millions of elements in 3D. There are some older papers claiming to use ``knowledge-based methods'' to generate meshes \cite{Manevitz02, Cinar02}, recognize model features \cite{Kim92, Wu18}, or even decompose geometric models \cite{Takata99,Lu99} but none of them used ML as we know it. Recent papers on CAD and ML have focused a lot on Shape Matching \cite{Qin14, Bronstein17, Boussuge19} and to a lesser extent on CAD model generation \cite{Wu21, Lin20} and cleanup \cite{Danglade14, Owen19}. 

\subsection{Our Approach}
This article presents a proof-of-principle demonstration of a novel AI-assisted method for decomposing complex geometric models into blocks by applying it to planar shapes with straight, axis-aligned edges. Our approach uses a reinforcement learning (RL) method to teach an agent a good sequence of steps to take in order to cut the input model into meshable blocks. In RL, an \textit{agent} learns by taking \textit{actions} in an \textit{environment} based on the \textit{state} of the environment. Each action moves the agent into a new state and grants it a \textit{reward}. With a targeted balance of exploration vs exploitation, the agent learns a \textit{policy} that maximizes the cumulative reward over a sequence of actions. RL closely mimics how human analysts learn to block decompose complex shapes and in recent years, RL, combined with deep neural networks (DNN), has matched or surpassed human-level skill in several fields \cite{mnih-2015-qlearn}. It is worth noting that this study is different from the use of reinforcement learning for image segmentation in medicine \cite{SahbaF2006ARLF} or in video processing \cite{Han_2018_CVPR, Wang_2020_CVPR} or segmentation of 3D point clouds \cite{tiator2020point}.

There are many challenges in applying reinforcement learning to the problem of block decomposition of complex geometric models.  Unlike common scenarios like learning to play a game or navigate a warehouse where the environment is fixed, \textit{our environment is dynamically changing} as we make cuts. Thus a naively formulated global observation set will vary in size as the episode progresses making it unsuitable for traditional neural networks. The agent itself has \textit{multiple types of decisions} to make - \textit{where to perform a modification} and \textit{what type of modification to make} (full cut, partial cut, etc.). Additionally, the \textit{parameters of the modification are continuous} (for example, the angle of a cut) and the agent must learn a distribution of expected rewards over the continuous parameter space. Finally, the task of the agent is not to master the decomposition of one particular geometric model - rather \textit{the ultimate goal is to learn a generalizable policy} that can be applied to new configurations. 

To tackle this problem, we set up an RL agent that takes an input geometric model that is planar with straight, axis-aligned edges. The agent recursively subdivides it into simpler parts using axis-aligned cuts. The environment is a custom setup that can read a geometric model and answer queries about it (e.g. how many vertices, how many edges connected to a vertex, the angle formed by two edges at a vertex). The agent also uses the capabilities of the geometric modeler to make modifications to the shape - in this particular study, the modification is slicing the geometric model into two or more pieces from a model vertex. The quality of the resulting parts (reduced complexity, low aspect ratio) determines the reward the agent receives. An episode ends when the input is decomposed into all quadrilateral blocks. In the results section, we demonstrate that our RL agent quickly learns which cuts to make and where to make them to maximize its rewards.

While the method is currently demonstrated on a simple problem that may be solved using procedural algorithms, it encapsulates most of the principles required to apply it more complex 2D and 3D shapes. We believe this is the first time such a reinforcement learning approach has been used to tackle the problem of block decomposition.

\section{Methodology}
\begin{figure}
    \centering
    \includegraphics[width=0.9\columnwidth]{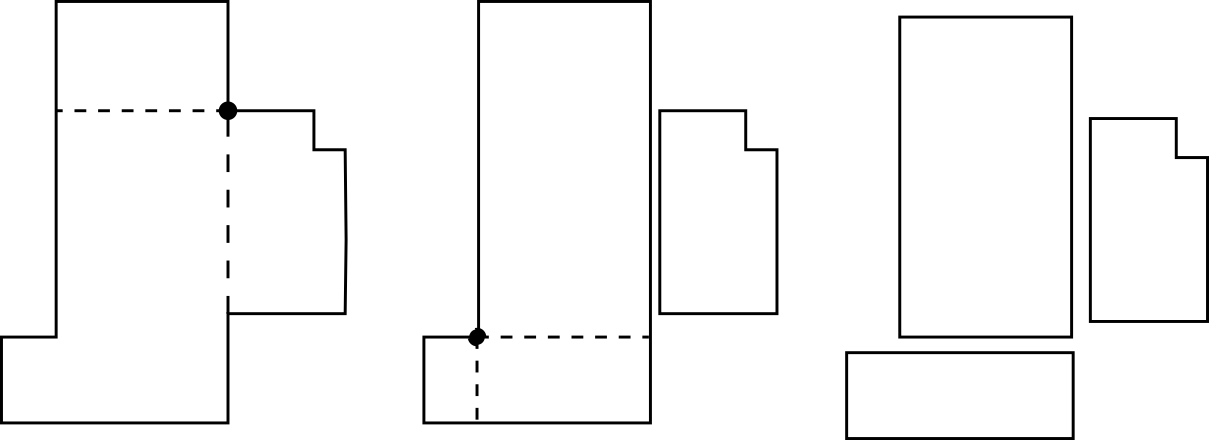}
    \caption{Recursive slicing of the model. The left figure shows the original model and a vertex at which the agent is poised to act along with the two cuts it can make. The middle figure shows the same process the two shapes from the first action and the choices for the next action. The right figure shows the shapes arising from the second cut.}
    \label{fig:recursive-cuts}
\end{figure}

We have developed a customized RL framework that learns how to effectively decompose geometric models into blocks by exploring the effect of different geometric model modifications. While most components of our RL framework are set up for general problems in 2D and 3D, this study is limited to decomposing planar shapes with straight, axis-aligned edges. The CAD model is described using a full-featured 3D geometric modeler called OpenCascade \cite{OpenCascade} but for the purposes of this discussion, it can be considered to be one or more planar shapes, each of which is described by a sequence of model vertices and model edges. During each step of the training phase, the agent picks a vertex of the geometric model, observes the state and makes a geometric modification. Currently, the only geometric modification the agent can make is a full cut, i.e., slice the geometric model into two or more parts using an infinite line (See Figure~\ref{fig:recursive-cuts}). While we use an RL technique that allows for a continuous action space (e.g. cuts originating at any location and angled arbitrarily), we restrict the cuts in this study to only originate from a model vertex and be aligned with the X- or Y-axis. Since the geometric model evolves as the agent makes cuts, the size of a global observation set for the full model, e.g. the list of vertices, also changes and cannot be used directly as input to a traditional neural network. Therefore, following the idea of Pan \cite{Pan22}, we have designed a fixed size local observation set at each model vertex to feed to the neural networks in the RL framework. The iterative application of this sequence of steps - select vertex, construct local observations, make a cut, evaluate the quality - allows the agent to learn to block decompose the geometric model. In order to learn a policy to efficiently perform such a decomposition, the agent is trained via feedback from the environment: cuts that produce a \emph{good partition}, e.g. resulting in quadrilateral blocks with good aspect ratios are \emph{rewarded}, while cuts that produce a \emph{bad partition}, e.g. high aspect ratios in its decomposed parts, high variance in the areas of its decomposed parts or cuts that do not affect the model (cutting along a side) are \emph{penalized}. The policy learned in this way can then be applied to perform block decomposition of other planar axis-aligned shapes.

\subsection{Soft Actor-Critic-based RL Architecture}
Our framework uses the soft actor-critic (SAC) reinforcement learning algorithm introduced in~\cite{haarnoja-2018-sac}. The SAC method provides a sample efficient (i.e. moderate data collection demands) and stable, model-free\footnote{Here we are talking of a model of a general environment for RL, not a geometric model.}, deep RL algorithm for continuous state and action spaces. While it may be argued that this problem might be tackled with a deep Q-network, the reason for using a SAC-type algorithm is to build a framework that can be generalized to more complex 2D and 3D models that require arbitrarily angled or partial cuts from any boundary location.

There are three main components in the SAC algorithm: 
\begin{enumerate}
    \item An actor-critic architecture with separate policy and value function networks,
    \item An off-policy formulation that enables reuse of previously collected data for efficiency, and
    \item Entropy maximization to enable stability and exploration. 
\end{enumerate}

The implementation of the soft actor-critic architecture includes three separate networks: an \emph{actor} network, a \emph{critic} network and a \emph{value function} network that are optimized jointly during training. As discussed by~\cite{haarnoja-2018-sac}, this not only provides flexibility to handle large continuous domains, but can also help to stabilize training.

\subsubsection{Actor Network}
The actor network outputs a probability distribution over the action space $\mathcal{A}$ and is also in charge of executing actions. In our case, it is implemented as a traditional neural network that receives as input a local observation (described below). Its output determines the probability for each of the two directions allowed for cuts from a given vertex: along the X-axis or Y-axis. Note that training uses a stochastic actor, where the selection of a cutting direction is made randomly weighted by the estimated probabilities, while, during deployment, the actor behaves deterministically selecting the action with the maximum estimated probability. The stochasticity is useful to maximize the entropy of the actor network and encourage exploration of the environment in the training phase.

\subsubsection{Critic Network}
The critic network qualifies how good the allowed actions are for a given state. It is similar to a Q-network in Deep-Q learning~\cite{mnih-2015-qlearn} in that it learns to approximate the Q-value of actions in a given state, i.e. it learns to approximate the reward for a given \emph{action-state pair}, along with all future rewards along the expected trajectory. In our case, it is also implemented as a traditional neural network that receives as input a local observation and determines the Q-value (quality) of X-axis and Y-axis cuts.

\subsubsection{Local Observation}
\begin{figure}
 \centering  
   \includegraphics[width=0.2\textwidth]{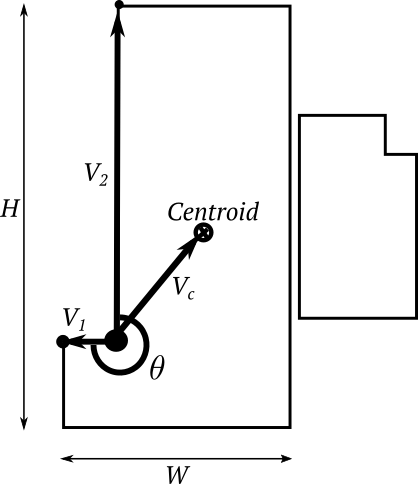}
   \caption[]{Features included in the local observation: Vectors to neighboring vertices ($V_1$, $V_2$) and vector to centroid of the shape ($V_c$), angle of the vertex corner ($\theta$), aspect ratio of full shape ($H/W$).}
    \label{fig:local_obs}
\end{figure}

The actor and critic networks are represented as traditional neural networks that expect a fixed input structure and, thus, are not able to handle the varying size and complexity of the evolving environment (i.e. the changing collection of vertices and edges as the geometric model is sliced repeatedly). Hence, we construct a special fixed structure to capture important \emph{local} shape information \emph{observed} at a chosen model vertex. The features included in this structure are: 
\begin{itemize}
    \item Vectors to the two neighboring vertices
    \item Type of interior angle formed by the two vectors (acute, right, obtuse, reentrant, etc)
    \item Vector to the centroid of the shape being processed
    \item Aspect ratio of the shape being processed
\end{itemize}
A schematic of the local observation features can be found in Figure~\ref{fig:local_obs}. As explained later, the complexity of observations at model vertices in our study remains fixed because the two parts resulting from a cut are treated as independent parts for the next cut - thus every model vertex remains connected to two vertices.

\subsubsection{Value Network}
\begin{figure}
  \centering
  \begin{subfigure}{.22\textwidth}
    \includegraphics[width=\textwidth]{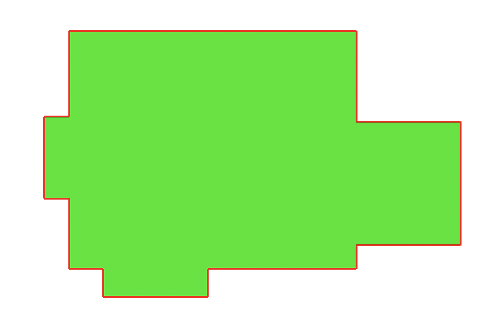}
    \subcaption{Example planar shape.}
    \label{fig:shape}
  \end{subfigure}%
  \hfill
  \begin{subfigure}{.2\textwidth}
    \includegraphics[width=\textwidth]{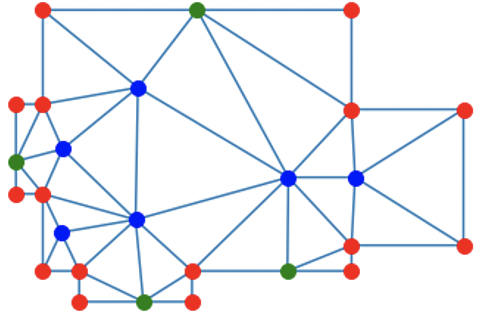}
    \subcaption[width=\textwidth]{Triangulation input.}
    \label{fig:gnn}
  \end{subfigure}
  \caption{Example planar shape and corresponding triangulation input to the value network. In the triangulation input, red mesh vertices lie on model vertices (vertex type 2), green mesh vertices lie on model edges (vertex type 1), blue mesh vertices lie in the interior (vertex type 0).}
  \label{fig:shape_gnn}
\end{figure}

The value network qualifies how good a particular state is. In other words, this network learns to approximate the expected reward and future rewards the actor will receive in a given state. In our case, this network allows the actor to choose the next vertex to perform a cut. Thus, it is more appropriate to regard this network as being able to approximate the expected reward and future rewards the actor will receive for making a cut from a specific vertex. For efficiently capturing all the relevant vertex-level information for the full model, this network must be able to handle the varying collection of vertices produced during shape decomposition. Hence, we implement this network as a graph neural network (GNN), specifically as a SplineCNN network~\cite{fey-2017-spline}. The network receives as input a triangular mesh of the planar model. We can control the resolution of this triangular mesh, usually preferring coarse meshes to avoid excessive computational burden. We tag the  mesh vertices as being coincident with model vertices, lying on a model edge or lying in the interior of the model as shown in Figure~\ref{fig:gnn}. Furthermore, notice that the GNN not only allows us to work with a changing number of vertices, it also enables the incorporation of spatial geometric information of the current decomposition state, information that would be much more difficult to encode using a traditional NN.

Although the value network produces an output at every mesh node, only the outputs at the model vertices (i.e. red points in Fig.~\ref{fig:gnn}) are considered. As stated above, the output value of the value network at a model vertex is an approximation to the expected reward and future rewards if a cut is made at that vertex. With this structure in place, we can chose a vertex to perform a cut at every step of an episode. Mimicking the stochastic actor concept, the set of values produced by the value network on the model vertices is used during training as probability weights and the vertex to perform a cut is randomly selected using these weights with the goal of encouraging exploration. In contrast, the selection is deterministic during deployment and the vertex with the highest output of the value network is selected to perform a cut.



\subsection{Off-Policy Formulation}
The SAC algorithm uses off-policy actor-critic training, combined with a stochastic actor as described before, which results in a more stable and scalable algorithm. Such a strategy allows it to reuse past experience to train the models and increases the sample efficiency. It is implemented by storing a distribution $\mathcal{D}$ of previously sampled states, actions and rewards, and using it as a replay buffer during training. We follow this approach during training which alternates between collecting experience from the environment by applying the current policy, and updating the networks via stochastic gradients computed from batches sampled from the replay buffer. 

\subsection{Entropy Maximization}
Unlike the regular actor-critic framework, SAC rewards entropy in its actions by optimizing policies to maximize both the expected return and the expected entropy of the policy. This encourages exploration of the environment and makes the algorithm more robust and capable of general learning, rather than just memorization. The maximum entropy policies are also robust to estimation errors and improve exploration by allowing the acquisition of diverse behaviors.

\subsection{Reward Function}
The reward function is a critical component of the RL framework and contributes to the effectiveness with which the agent carries out the task at hand. In our case, we devise a reward function to
\begin{itemize}
    \item Encourage creating quadrilateral parts
    \item Discourage cuts that do not affect the geometric model (e.g. cutting along a side)
    \item Discourage high variance in the areas of its decomposed parts
    \item Discourage high aspect ratios in its decomposed parts
\end{itemize}

Once the geometric model is fully decomposed into blocks, the agent gets a bonus reward and the episode concludes. The exact form of the rewards used for this study are given in the results section.

\subsection{Training Phase}
The training phase is composed of a collection of episodes, each \emph{episode} consisting of all the steps needed for decomposing a given geometric model. During a training episode, the agent uses the value network output to select a vertex to cut, and the actor network output to select the particular action to take. Both of these are done stochastically to ensure a higher level of exploration during training. 

The steps listed below are iterated during a training episode

\begin{enumerate}
\item Triangulate the shape being processed
\item Run the value network on the triangulation to generate weights at \textit{mesh} vertices 
\item \textit{Stochastically} select a \textit{model} vertex based on value network outputs 
\item Compile a local observation at the vertex
\item \textit{Stochastically} choose a direction for a cut at the vertex based on actor network probability outputs
\item Split the geometric model into two or more parts along the chosen direction
\item Compute the new state and reward
\item Store sampled states, actions and rewards in the replay buffer
\item Update parameters for every network following the gradient step
\item Pick another non-quadrilateral part from the geometric model decomposition and repeat from step 1
\end{enumerate}

Geometric models are loaded repeatedly from the training set, one per episode. A set number of episodes is run during training. The training of all the networks uses the Adam optimization algorithm. The functions optimized in each case are the same as in the SAC original work. There is, however, a slight difference in the value network: when calculating the loss, the network only propagates loss for the node that was chosen to make a cut from.
 
Note that a cut goes fully through the shape and splits it into two or more parts (see Fig.~\ref{fig:recursive-cuts}). Instead of keeping the model as a collection of generated parts, we treat each part as a new shape to explore. Thus at each step we split the model, set aside quadrilateral parts, and put the remaining parts in a processing queue. This approach sacrifices the full model view, but makes it simpler and more robust since the agent does not encounter a local state of ever increasing complexity and there is no need to accumulate the knowledge of how the parts build up. An additional benefit of this approach is that each new part generated is a training data sample for the agent.

\subsection{Deploying the Trained Framework}

After the framework is trained, the combination of value network and actor network constitute the learned policy. The decomposition of new geometric model proceeds as follows (with similarities to the training phase):
\begin{enumerate}
\item Triangulate the shape being processed
\item Run the value network on the triangulation to generate weights at mesh vertices 
\item \textit{Deterministically} select a model vertex with highest value output by the value network
\item Compile a local observation at the vertex
\item \textit{Deterministically} choose the cut direction with the highest probability as predicted by the actor network
\item Split the geometric model into two or more parts along the chosen direction
\item Compute the new state and reward
\item Pick another non-quadrilateral part from the geometric model decomposition and repeat from step 1
\end{enumerate}

Crucially, at the end of the decomposition, all the shapes are merged backed together while retaining the boundaries between them. Thus vertices that appear on the boundary of one block are also reflected in the boundary of adjacent blocks. The merged model is then meshed using well-known procedures. In our case, we import the parts into the CUBIT geometric modeling and meshing package \cite{CUBIT}, use its imprint-and-merge functionality to recreate a single geometric model (with internal cuts) and apply a mapped meshing algorithm.

\section{Numerical Experiments}

\subsection{Data Sets}
\begin{figure}
 \centering
 \begin{subfigure}{.5\textwidth}
   \includegraphics[trim=0 0 0 27,clip, width=0.9\textwidth]{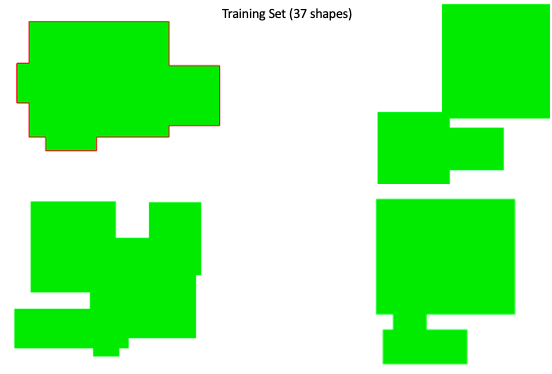}
   \subcaption[]{}
   \label{fig:trdata}
 \end{subfigure}
 \begin{subfigure}{.5\textwidth}
     \includegraphics[trim=0 0 0 25,clip, width=0.9\textwidth]{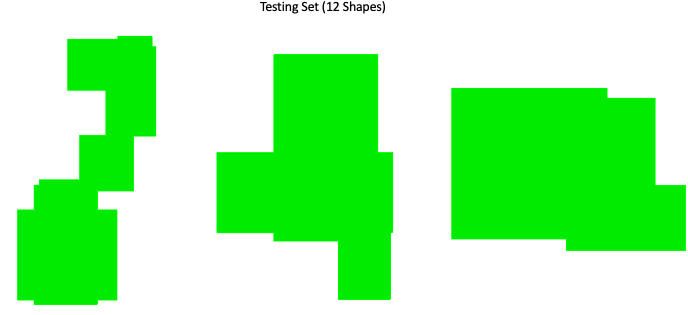}
     \subcaption[]{}
     \label{fig:testdata}
 \end{subfigure}
 \caption{Samples from the (a) training data set containing 37 models and (b) test data set containing 12 models.}
\end{figure}

Our training and testing data set includes 49 planar shapes with straight, axis-aligned edges. These were generated using our python script that invokes the CUBIT package \cite{CUBIT} to randomly generate and combine 2 to 10 rectangles. The training and test data sets consist of 37 models and 12 independent models respectively (Figures \ref{fig:trdata}, \ref{fig:testdata}).

\subsection{Network Architecture}

All networks are implemented using PyTorch~\cite{paszke-2019-pytorch} and PyTorch Geometric~\cite{fey-2019-pyg}. The architectures used are described next.

\subsubsection{Actor and Critic Networks}
These networks are traditional feed-forward NN composed of 4 fully-connected layers, with 256, 128, 64 and 2 neurons, respectively (the last of these layers is the output layer). We use rectified linear unit (ReLU) activation functions after each of these layers, except for the last layer in the critic network\footnote{We actually use two equally parameterized critic networks as in the original SAC work, see~\cite{haarnoja-2018-sac} for details.} that uses a linear activation function. The input dimension is 9, corresponding to the size of the local observation: 2 dimensional (2D) vector for each of the 2 neighboring vertices, 2D vector to centroid, 1 value for angle at vertex and 2 components to represent the aspect ratio. The networks have 2 outputs which correspond to the dimension of the action space (i.e. 2 cut directions: X-axis or Y-axis).

\subsubsection{Value Network}
This network is a GNN. It contains 1 SplineCNN layer, followed by 7 residual blocks and 1 final SplineCNN output layer. Each residual block is composed of 2 SplineCNN layers. There are batch normalization layers after all the SplineCNN layers except the output layer. We use exponential linear unit (ELU) activation functions except in the output layer. Every SplineCNN layer has a kernel size of 5, meaning the 2D B-spline function for the continuous kernel has 25 defining points, with 5 points on each axis. The number of nodes in the graph is arbitrary and depends on the triangulation of the shape. Each node in the graph input layer has 3 features because each node has one-hot encoded vector features: $(1, 0, 0)$ represents interior point, $(0, 1, 0)$ represents boundary point and $(0, 0, 1)$ represents model vertex. Each node in the graph output layer has 1 feature corresponding to the value function for that node, but only nodes corresponding to model vertices are taken into account. The first SplineCNN layer has 64 features. The residual blocks have 128, 256, 128, 64, 32, 16 and 8 features, respectively. Note that if the number of features does not change through a residual block, the input features to the residual block are simply summed with the output features. However, if the number of features does change through a block, the skip connection contains 1 SplineCNN layer, with as many features as the features in the block. All our residual blocks change the number of features.

\subsubsection{Reward Function}
Assume that a splitting action on a shape results in $N$ new shapes, with $N_q$ of them being quadrilaterals. Let the areas of the shapes be $A_i,\; i=1,N$, and aspect ratios $R_i,\; i = 1,N$ (where the aspect ratio of a shape is defined as the ratio of the longest side to the shortest side of its bounding box). Also, let the average area of all the shapes be $\bar{A}$. Finally, define a penalty function $p(N)$ which is 1 when $N=1$ (no new shapes are created by an action) and 0 otherwise.

The reward $\mathcal{R}$ is defined as
\begin{equation}
\small
    3\Biggl[\Biggl(\frac{N}{\sum_i{R_i^2}}\Biggr)^\frac{1}{2} - \Biggl(\frac{\sum_i{(A_i-\bar{A})^2}}{\sum_i{A_i}}\Biggr)^\frac{1}{2}\Biggr] + 10\frac{N_q}{N} - 5p(N) - 1
\end{equation}

Note that minimum possible aspect ratio is 1 and therefore the leading term takes a maximum value of 1 when all shapes are squares. The second term which measures variance in the areas of the shapes takes a minimum value of 0 when all the areas are equal. The third term is a maximum if the action results in all quads ($N_q = N$). The fourth term serves is a penalty for actions that result in no new shapes (i.e. some cuts along edges). Thus the maximum reward is obtained when the action cuts the shape into all squares of equal area.

\subsection{Testing and Reward Convergence}
As the model learns using the training set, the RL framework's learning is periodically checked against the test set. In a testing episode, the vertex at which to act and the action to take are chosen deterministically to maximize the reward - a vertex with the highest output from the value network is chosen, and the action with the highest probability from the actor network is applied at that vertex.

\begin{figure}
  \centering
  \begin{subfigure}{.5\textwidth}
    \includegraphics[width=\textwidth]{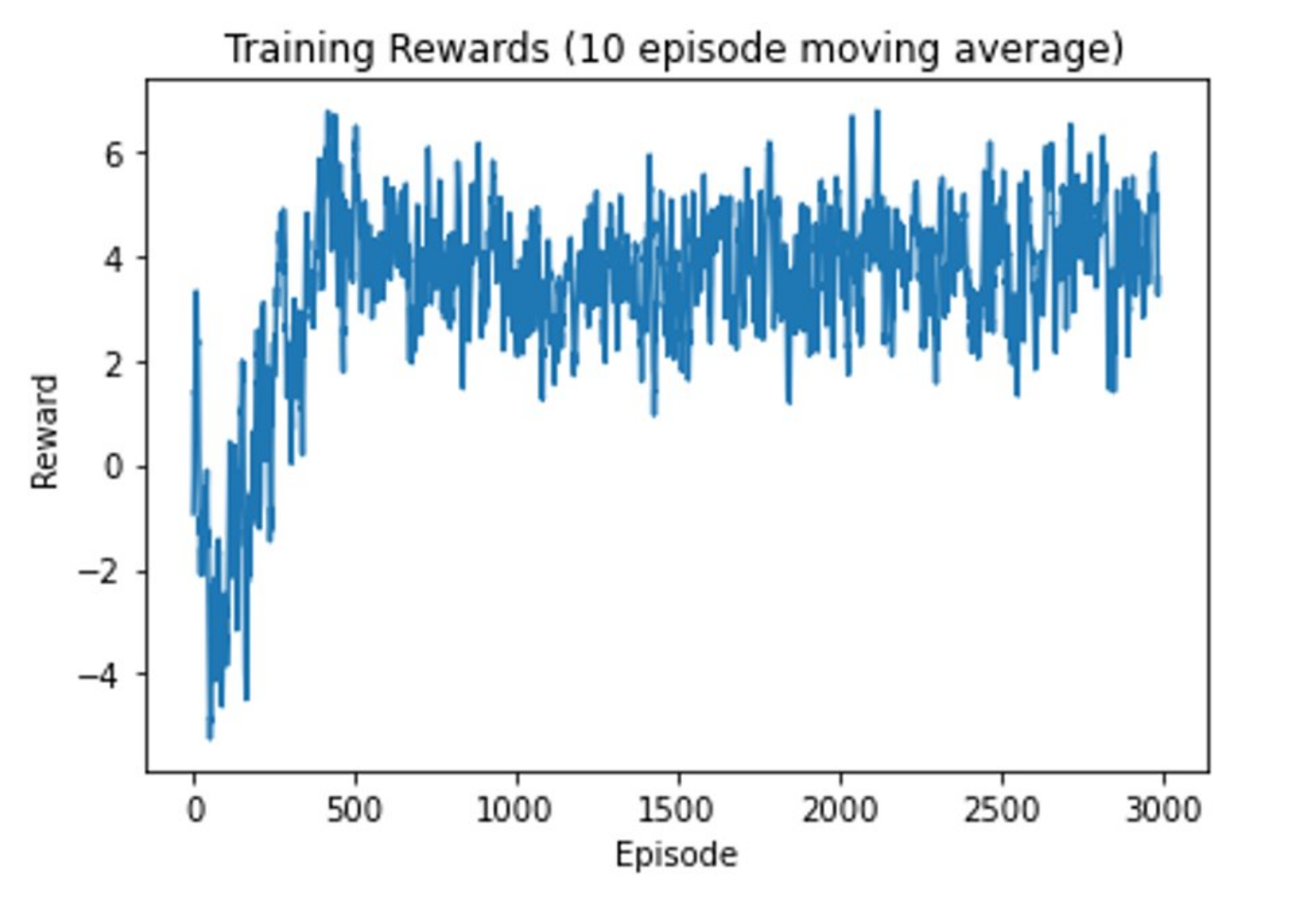}
    \subcaption[]{Rewards obtained by the agent on the training dataset as it trained.}
    \label{fig:rc1}
  \end{subfigure}%
  \hfill
  \begin{subfigure}{.5\textwidth}
    \includegraphics[width=\textwidth]{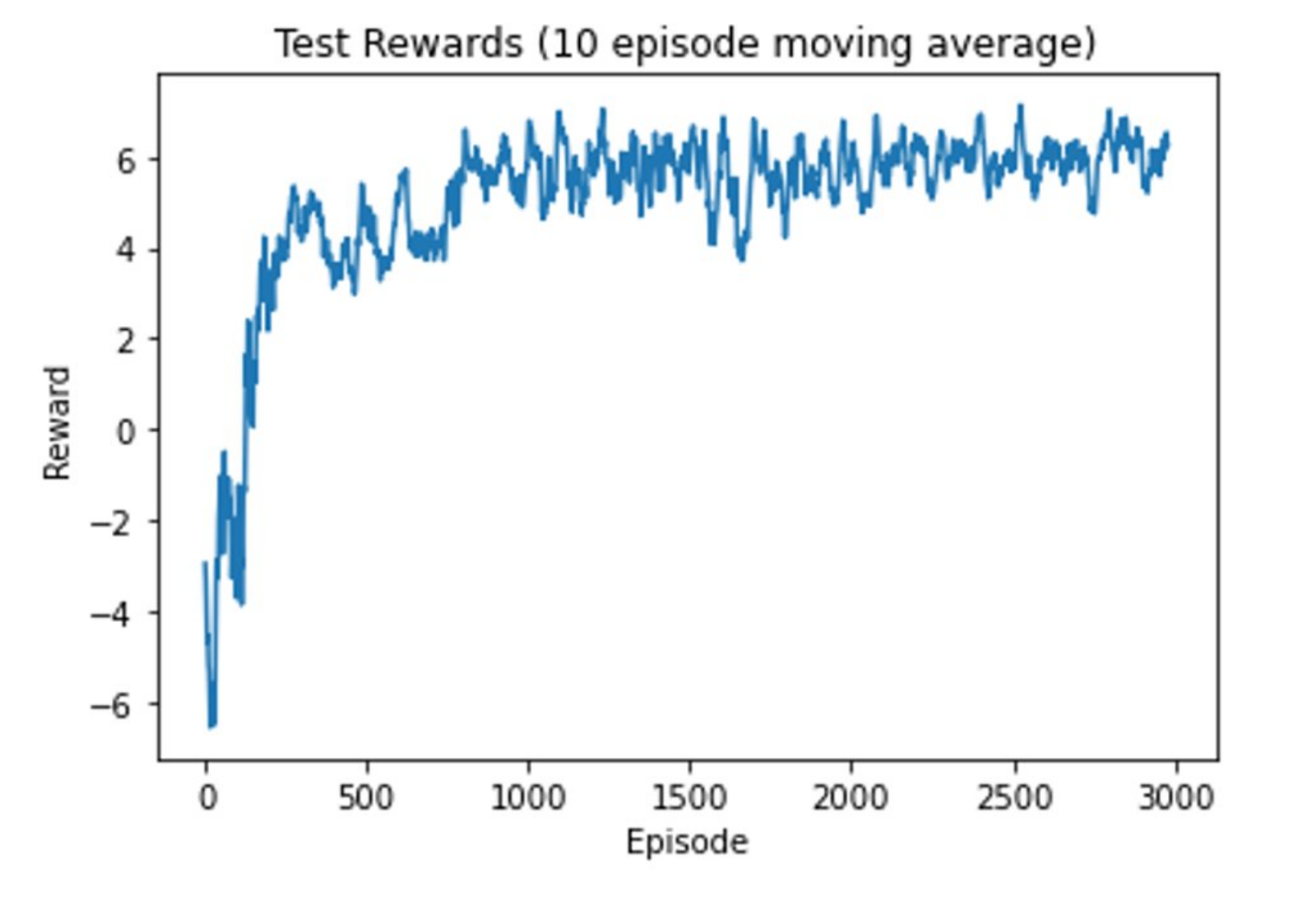}
    \subcaption[width=\textwidth]{Rewards obtained by the agent on the test dataset as it trained.}
    \label{fig:rc2}
  \end{subfigure}
  \caption{Reward convergence obtained by the training.}
\end{figure}

Figure \ref{fig:rc1} shows that a moving average of rewards (over 10 episodes) obtained by the RL framework during the training phase. Figure \ref{fig:rc2} shows the convergence of a moving average of rewards during the periodic testing episodes. We observe that after only around 1500 episodes of training (an hour or so of training time) the model learned to obtain consistently high rewards on its training set, but also on the test set of shapes it has never trained on. The oscillations in the reward plot of the training set indicate that the agent is continuing to favor exploration rather than exploitation. The good reward convergence seen on the test set implies that the agent is steadily gathering generalizable knowledge about the decomposition problem for this category of shapes.
        
\subsection{Decomposition Examples}
\begin{figure}
 \centering  
   \includegraphics[width=0.5\textwidth]{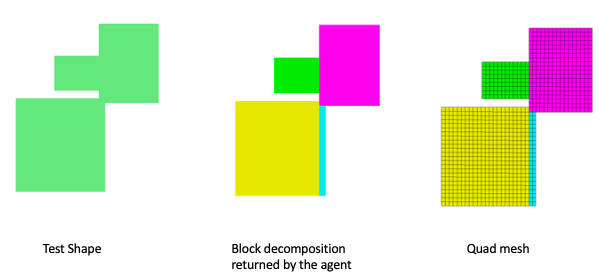}
   \caption[]{The block decomposition (middle) returned by the agent for the test shape shown on the left and its mesh from CUBIT (right).}
    \label{fig:dex1}
\end{figure}

\begin{figure}
  \centering
  \includegraphics[width=0.5\textwidth]{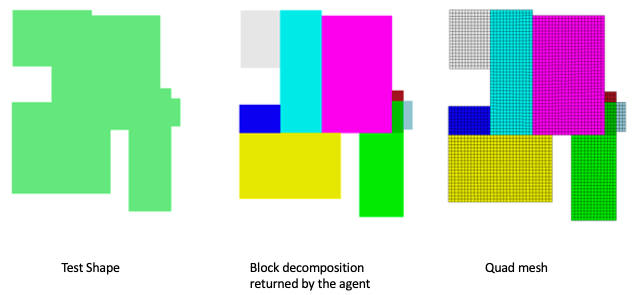}
  \caption{The block decomposition (middle) returned by the agent for the test shape shown on the left and its mesh from CUBIT (right).}
   \label{fig:dex2}
\end{figure}

Finally, in Figures \ref{fig:dex1} and \ref{fig:dex2}, we present two examples of block decompositions obtained for test shapes (i.e. shapes that the agent never trained on). It showcases the learned knowledge of the agent after it was trained. The block decompositions were then meshed using CUBIT to generate quadrilateral meshes of the decomposed shape. 

\section{Conclusions}
We have demonstrated a novel reinforcement learning-based AI method to decompose input CAD shapes into well shaped blocks that can be meshed for numerical simulations. The results show that an agent using the SAC reinforcement learning framework can learn a block decomposition policy that generalizes to new planar, axis-aligned shapes.

While this proof-of-principle demonstration is restricted to simple 2D shapes and elementary geometric model modifications, it contains most of the principles required to generalize it to more complex shapes in 2D and 3D. The environment is based on geometric modelers which regularly handle complex 3D shapes with curved boundaries. The agent's actions are modeled on the types of operations a human agent decomposing a shape will execute using a geometric modeler (e.g. planar model cuts). The use of Soft-Actor-Critic framework allows for continuous actions (e.g. cuts at an angle) in the future. Similarly, the rewards are based on the quality evaluation of the blocks used by meshing algorithms and analysts. The issue of variability in the starting environment and the dynamic evolution of the environment are already addressed in this simple problem using a graph-based value neural network. Thus, we can reasonably surmise that the method can eventually be generalized to address the real problem of decomposing 3D shapes thereby alleviating one of the long standing problems in meshing.

In the future, we will expand this research to tackle more complex 2D and 3D shapes. Tackling non-axis aligned 2D shapes will require continuous actions like cuts at arbitrary angles. Expanding to more complex curved geometric models will require expansion of the types of actions to include partial cuts or some other templated subdivision (like making a square internal boundary inside a circular part). The reward function definitions may also have to be refined further. Expanding the method to 3D requires tetrahedral meshes for the value network, an expanded set of observations, generalized reward functions and more types of geometric modifications. 

\section{Acknowledgements}
This work was supported by the U.S. Department of Energy for Los Alamos National Laboratory {\small (LANL)} under contract\\ \textsf{\small 89233218CNA000001}. This publication is approved for release as LANL report number LA-UR-22-29605.

\bibliography{bibfile1}

\end{document}